\documentclass[11pt,onecolumn]{IEEEtran}
\usepackage{hyperref}       % hyperlinks
\usepackage{url}            % simple URL typesetting
\usepackage{booktabs}       % professional-quality tables
\usepackage{amsfonts}       % blackboard math symbols
\usepackage{nicefrac}       % compact symbols for 1/2, etc.
\usepackage{microtype}      % microtypography
\usepackage{amsmath}
\usepackage{amsthm}
\usepackage{multirow}
\usepackage{amssymb}
\usepackage{bm} 
\usepackage{graphicx}
\usepackage{subfigure}
\usepackage{algorithm}
\usepackage[noend]{algpseudocode}
\usepackage{colortbl,booktabs}
\usepackage[margin=1.9cm]{geometry}
\usepackage{xcolor}
\usepackage{etoolbox}
\definecolor{mygray}{gray}{.9}

\newtheorem{remark}{Remark}

\theoremstyle{definition}

\usepackage{amsmath}
\DeclareMathOperator*{\argmax}{arg\,max}

\begin{document}
\title{Demystifying CNNs for Images by Matched Filters}
\author{Shengxi Li, Xinyi Zhao, Ljubisa Stankovic, Danilo Mandic, 
\thanks{Shengxi Li (ShengxiLi2014@gmail.com) is with the Beihang University and Imperial College London; Xinyi Zhao (zhaoxy\_@buaa.edu.cn) is with the Beihang University; Ljubisa Stankovic (ljubisa@ac.me) is with the University of Montenegro; Danilo Mandic (d.mandic@imperial.ac.uk) is with the Imperial College London. }}
\maketitle
\begin{abstract}
The success of convolution neural networks (CNN) has been revolutionising the way we approach and use intelligent machines in the Big Data era. Despite success, CNNs have been consistently put under scrutiny owing to their \textit{black-box} nature, an \textit{ad hoc} manner of their construction, together with the lack of theoretical support and physical meanings of their operation. This has been prohibitive to both the quantitative and qualitative understanding of CNNs, and their application in more sensitive areas such as AI for health. We set out to address these issues, and in this way demystify the operation of CNNs, by employing the perspective of matched filtering. We first illuminate that the convolution operation, the very core of CNNs, represents a matched filter which aims to identify the presence of features in input data. This  then serves as a vehicle to interpret the convolution-activation-pooling chain in CNNs under the theoretical umbrella of matched filtering, a common operation in signal processing. We further provide extensive examples and experiments to illustrate this connection, whereby the learning in CNNs is shown to also perform matched filtering, which further sheds light onto physical meaning of learnt parameters and layers. It is our hope that this material will provide new insights into the understanding, constructing and analysing of CNNs, as well as paving the way for developing new methods and architectures of CNNs. 
\end{abstract}

\section{Introduction}
With the emergence of the ``Big data'' era, we are witnessing exponentially exploding amounts of information which continually arises from a variety of services and sources. The convenience of visualizing the world we live in has popularized images and videos as a preferred format in different multimedia applications, including social media, video conferencing and meta universe. Supported by the progress in sensors which are of continually broadened bandwidth, the high definition (HD) camera is now  the ``must-have'' in portable devices,  including intelligent mobile phones, laptops, and tablets, to name but a few. As such, images and videos represent the majority of multimedia content and account for more than 80\% data volume over the Internet \cite{cisco2022}. It is, however, already intractable for us to deal with such a large amount of image/video content in the wild; indeed, it would take one person approximately a century to watch over the videos uploaded to the Internet per day \cite{howmanyvideos2022}. It is therefore a necessity for us to employ intelligent machines to process images/videos with improved efficiency; these are designed with the aim to mimic human cognition in order to automatically extract useful information. 

An important family of intelligent machines are the artificial neural networks (ANNs) that learn from examples to exhibit similar behaviours to humans, i.e., through ground-truth labelled supervision \cite{artificial1988Hopfield,mandic2001recurrent}. Early ANNs employed fully connected (FC) layers whereby the input is linearly combined by means of a trainable  weight matrix. However, the FC neural networks quickly reach their limitations when processing large amounts of data, because the size of learnable parameters in the FC layers increases exponentially with the size of images and videos, a phenomenon known as the \textit{Curse of Dimensionality} \cite{bellman1966dynamic}. Since ANNs  need prohibitively large amounts of data to be appropriately trained and to reach acceptable performance, this issue significantly impedes the application of ANNs in current Big Data scenarios. 

Given that ANNs inherit their name from the neural networks in our brains, strategies of mimicking human brains therefore play an important role in designing ANNs, especially when handling image/video tasks. Indeed,  humans easily handle the excessive amounts of images and videos; human eyes behave equally to HD cameras of more than 100 million pixels, whereas the bandwidth between eyes and brains is only up to 8 Mbps \cite{koch2006much}, far below the processing speed of computers. One of the key reasons is the fovea mechanism in human vision system (HVS), where people perceive clearly a small region of 2-5$^\circ$ visual angle when viewing images and leave the other regions blurred \cite{simoncelli1996foundations}. More importantly, HVS processes visual information in a hierarchical manner, whereby the information is gradually extracted from low-level cues to high-level semantics by different functional cells \cite{hubel1962receptive}.  

These desirable features of HVS motivate the kernel design within convolutional neural networks (CNNs) \cite{lecun1989backpropagation,hinton2006fast}, and also guarantee their high efficiency when learning from large amount of images \cite{krizhevsky2017imagenet}. More specifically, instead of treating the input images on the whole, as in the case of ANNs, CNNs employ small trainable kernels to locally match representative patterns in an image, the same principle as with the fovea in HVS that understands an image through a hierarchical convolution of several representative regions. This strategy effectively reduces the amount of parameters in neural networks, thus allowing for stacking multiple layers to compose deep CNNs. The behaviour of deeply stacked layers in CNNs exactly mimics the hierarchical nature of HVS, and gradually extracts abstract features in a layer by layer manner, e.g., from edges to objects. This has made it possible for deep CNNs to achieve breakthroughs in manifold fields, for example, by beating humans in classifying images (ImageNet dataset \cite{he2016deep}) and winning against the best human player in the Go game \cite{silver2016mastering}.

Despite the success of CNNs, there are severe issues related to the inherently \textit{ad hoc} settings in their designs, that is, their \textit{black-box} nature, which prevent theoretical analysis and any associated physical meaning of the layers and parameters of learnt CNNs. This issue is especially prominent in deep CNNs for large-scale datasets. The basic principle of CNN is that each kernel aims to detect one local pattern which is invariant to translation, so that kernels in the  consecutive layers are responsible for hierarchical patterns at progressive stages; this is achieved by sliding kernels across the whole input images or features, i.e., the convolution operation. As its name implies, the convolution plays a key role in CNNs, however, the rationale for using the convolution operation still needs further theoretical justification. On the other hand, the convolution operator has been widely applied in the signal processing community and is supported by  theoretical analysis and know-how, together with its closely related matched filtering technique, which is continuously employed in radar, sonar and communications. It is therefore beneficial to aim to demystify CNNs from the matched signal filtering perspective, the subject of this tutorial. More specifically, we first introduce the matched signal filtering approach to find patterns in data, and proceed to focus on explaining several key principles of matching signals by given filters, that is, to detect patterns in noisy signals. Next, we revisit CNNs from the perspective of matched filtering, and show that the core convolution-activation-pooling chain in CNNs is indeed rooted in matched signal processing. We finally promote optimality in CNNs, through an adaptive method for matching patterns. It is our hope that this tutorial will pave the way for a seamless link between CNNs and signal processing techniques, which consequently may motivate new models for improving intelligent machines. 

\section{Matched Filtering for Signals}
It is oftentimes important to detect certain wavelets or patterns within (noisy) signals, for example, when recognising objects in radar signals or measuring similarities of speech signals. To this end, matched filtering employs a pre-defined template, in order to detect pre-defined patterns in the signals through the maximisation of the signal-to-noise ratio (SNR). More specifically, a filtering system, $h(\mathbf{t})$, represents basically the convolution operation with the signal, ${s}(\mathbf{t})$, given by
\begin{equation}\label{eq_system}
	g(\mathbf{t}) = (s*h)(\mathbf{t}) = \int s(\bm{\tau})\cdot h(\mathbf{t}-\bm{\tau})d\bm{\tau},
\end{equation}
where $*$ denotes the convolution operator. More importantly, when the signal ${s}(\mathbf{t})$ is assumed to be contaminated by additive noise, $\epsilon(\mathbf{t})$, that is, ${s}(\mathbf{t})={f}(\mathbf{t})+\epsilon(\mathbf{t})$, the matched filter is capable of detecting the pattern ${f}(\mathbf{t})$, through an optimal filter that maximises the SNR of the output signal, $g(\mathbf{t})$, against noise \cite{turin1960introduction}. The impulse response of the matched filter, $h(\mathbf{t})$, then simply represents a time/space reverse of the pattern, $f(\mathbf{t})$, to be detected, namely 
\begin{equation}\label{eq_matched_filter_design}
	{h}(\mathbf{t}) = a\cdot {f}(\mathbf{t}_0-\mathbf{t}), 
\end{equation}
where $a$ is a scaling constant that is related to noise spectrum and signal power, and $\mathbf{t}_0$ is a time delay. Without loss of generality, for notation convenience we set $a=1$ and $\mathbf{t}_0=\mathbf{0}$, such that ${h}(\mathbf{t}) = {f}(-\mathbf{t})$. In this way, the matched filtering operation can be formulated by combining \eqref{eq_system} and \eqref{eq_matched_filter_design}, to yield
\begin{equation}\label{eq_matched_filter_cont}
	g(\mathbf{t}) = (s*h)(\mathbf{t}) = \int s(\bm{\tau})\cdot f(\bm{\tau}-\mathbf{t})d\bm{\tau} = \int s(\bm{\tau}+\mathbf{t})\cdot f(\bm{\tau})d\bm{\tau}.
\end{equation}

\begin{remark}
Since the matched filtering operation achieves the best SNR performance,  it is evident from \eqref{eq_matched_filter_cont} that by sliding the time reversed template, $f(\mathbf{t})$, along the signal, $s(\mathbf{t})$, the convolution operation an equivalent way to detect patterns in noisy signals.	
\end{remark}

\begin{figure}
	\centering
	\subfigure[Matched filtering by convolution in 1D scenario]{\includegraphics[width=0.6\textwidth]{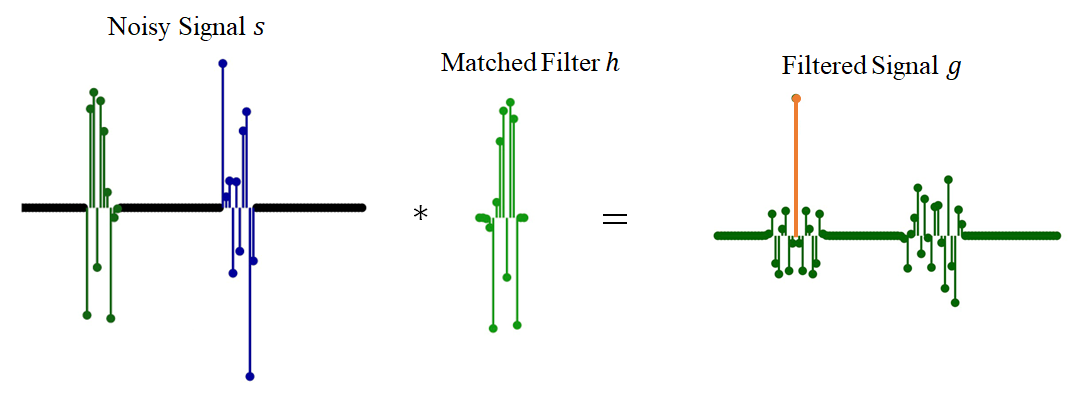}}
	\subfigure[Matched filtering by convolution in 2D scenario]{\includegraphics[width=0.6\textwidth]{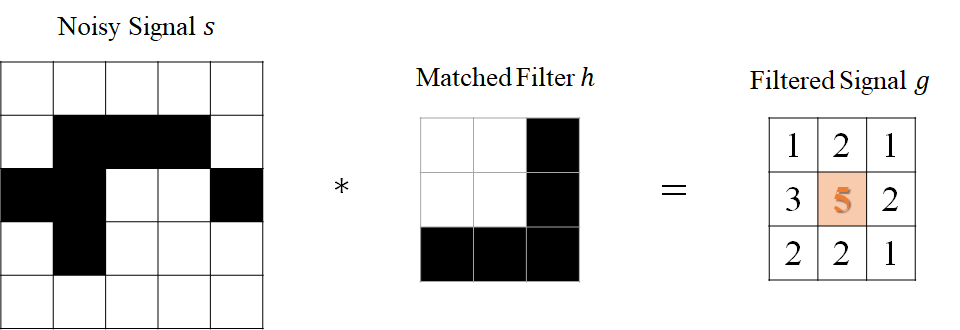}}
	\caption{Examples of matched filtering implemented through convolutions for detecting patterns in 1D and 2D noisy signals. The patterns of interest are both successfully detected by matched filtering operations.}\label{fig_single_filters}
\end{figure}

\begin{figure}
	\centering
	%\subfigure[Matched filtering operation]{\includegraphics[width=0.85\textwidth]{fig_matched_filtering_illustrate.png}}
	\subfigure[Convolution operation]{\includegraphics[width=0.85\textwidth]{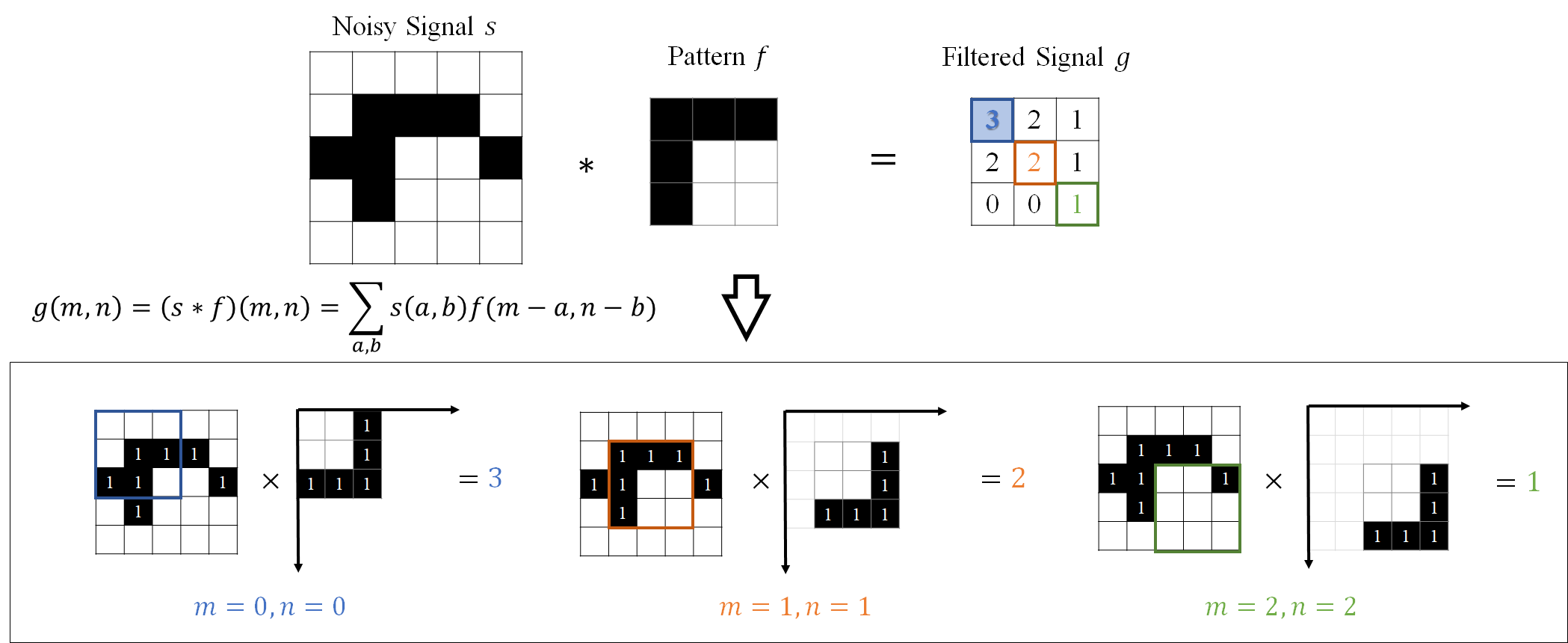}}
	\subfigure[Cross-correlation (matched filtering) operation]{\includegraphics[width=0.85\textwidth]{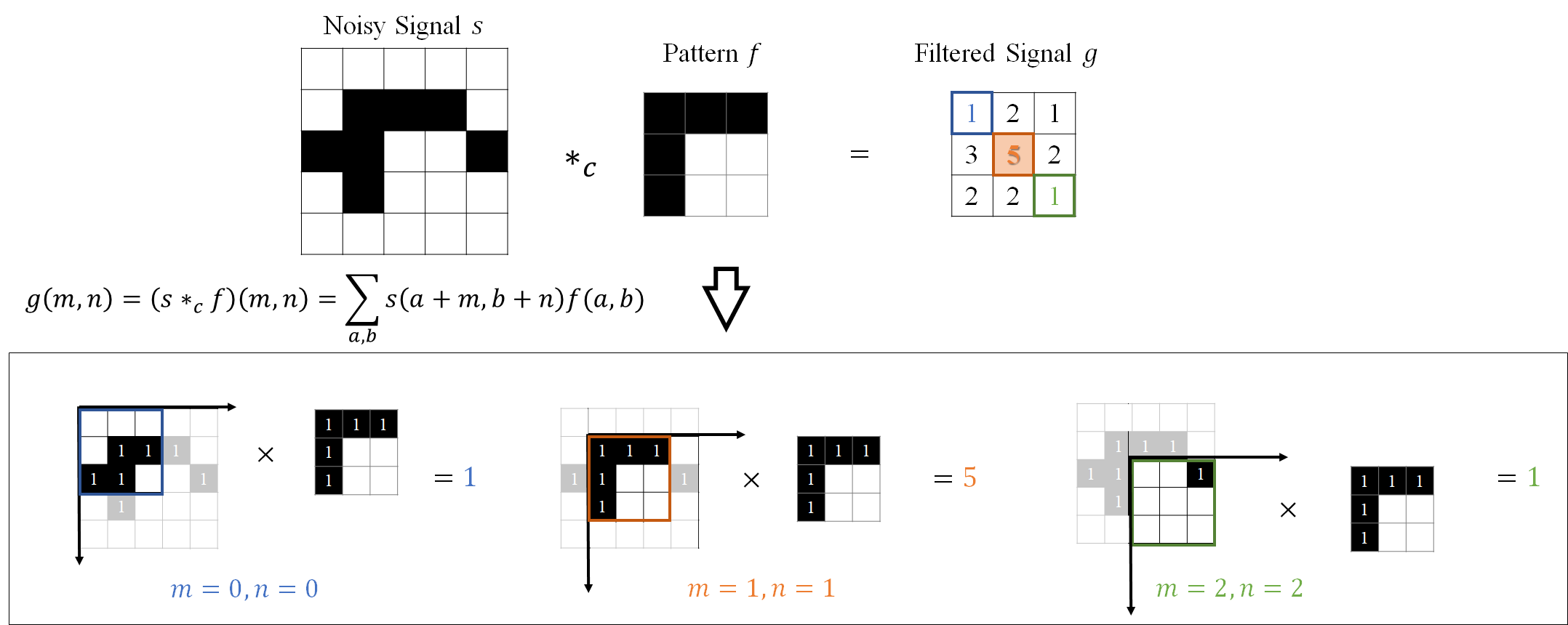}}
	\caption{Comparisons among the convolution (a) and cross-correlation (b) operations. The example aims at detecting an \textit{L} shape pattern in a noisy image. Note that the axis origin $(0,0)$ is the top-left point of the noisy signal, without any paddings. The grid with black colour represents the signal value of $1$, and white colour designates the value of $0$. The filtered signals, $g(m,n)$ at three different locations are given in the second panel of each subfigure, that is, for $\{m=0, n=0\}$, $\{m=1, n=1\}$ and $\{m=2,n=2\}$, as in \eqref{eq_matched_filter_disc}. Note that when the matched filter is asymmetric, filtering by the cross-correlation operation is different from the convolution. This way, the pattern of interest is successfully detected by maximising the filtered signal in (b), which is at the centre of the image, with the maximum value of $5$. However, the convolution operation fails to detect the pattern in (a). }\label{fig_single_asymfilters}
	%The detection process is performed by the cross-correlation between the noisy signal $s(m,n)$ and the matched filter $h(m,n)$, which is equivalent to the convolution with a temporal/spatial reversed version of the matched filter $h(-m,-n)$. }
\end{figure}
In this tutorial, we focus on the 2D convolution operation -- the main operator in CNNs, whereby the delay $\mathbf{t}$ lives in  2 dimensions, i.e., $(m, n)$; extensions to other dimensions are straightforward, based on this tutorial. We recommend our sister papers addressing the 1D and irregular matched filtering \cite{stankovic2021understanding,stankovic2021convolutional}. Upon discretising \eqref{eq_matched_filter_cont} in the 2D scenario, we arrive at 
\begin{equation}\label{eq_matched_filter_disc}
	g(m, n) = (s*h)(m,n) = \sum_{a,b} s(a,b) \cdot h(m-a, n-b) = \sum_{a,b} s(a+m,b+n) \cdot f(a, b).
\end{equation}
Typical choices of $(m,n)$ when training CNNs are the vertical and horizontal axis (locations) in an image. Fig. \ref{fig_single_filters} illustrates the matched filtering process in 1D and 2D scenarios. It is important to mention that matched filtering is invariant to the location shift due to the sliding mode of operation in convolution, whereby the identified pattern can be located by $\arg\max_{\{m,n\}} g(m,n)$. 

\begin{remark}
	In signal processing, the convolution of a reverse of a signal $f(m, n)$ is also called cross-correlation, which is a widely applied similarity measure between two signals. We denote the cross-correlation of $s(m,n)$ and $f(m,n)$ by $*_c$, so that
	\begin{equation}
		g(m, n) = (s*h)(m,n) = \sum_{a,b} s(a,b) \cdot h(m-a, n-b) = \sum_{a,b} s(a+m,b+n) \cdot f(a, b) = (s*_c f)(m,n).
	\end{equation}
In other words, the detection process is performed by the cross-correlation between the noisy signal $s(m,n)$ and the matched filter $h(m,n)$, which is equivalent to the convolution with a temporal/spatial reversed version of the matched filter $h(-m,-n)$. A comparison between the matched filtering, convolution and cross-correlation operations is illustrated in Fig. \ref{fig_single_asymfilters}.
\end{remark}

\begin{figure}
\centering
\includegraphics[width=0.9\textwidth]{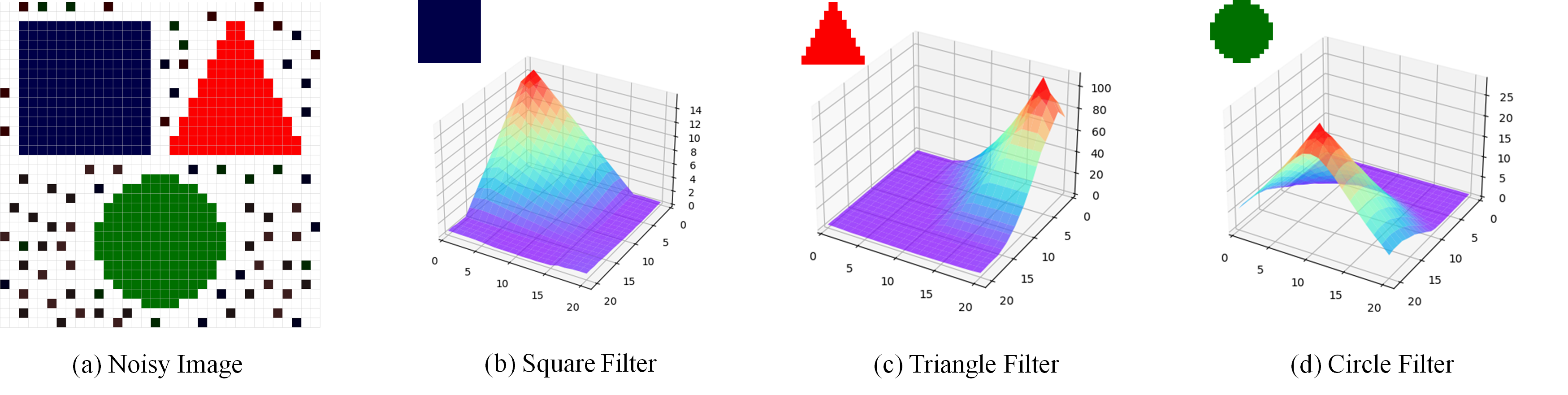}
\vspace{-.5em}
\caption{Illustration of detecting multiple patterns by matched filtering, that is, the blue square, red triangle and green circle  in a noisy image, given in (a). The dots in the image denote the additive noise. Images (signals) which are filtered by the square, triangle and circle matched filters (templates) are shown in (b), (c) and (d), respectively. Note that the matched filter successfully detects the correct pattern through the maximum of the cross-correlation, whilst suppressing the responses of other patterns and noise. This also explains the reason why the cross-correlation operation of matched filters is capable of achieving the best SNR performance against noisy signals. }\label{fig_multiple_filters}
\end{figure}

Furthermore, when aiming to detect multiple patterns, we can employ a set of matched filters $\{h_k(m,n)\}_{k=1}^K$ and decide the ``best'' matching by
\begin{equation}\label{eq_matched_filter_multiple}
	\{m_k, n_k\} = \argmax_{m,n} \{(s* h_1)(m,n), (s* h_2)(m,n), \ldots, (s* h_K)(m,n)\}.
\end{equation}
Fig. \ref{fig_multiple_filters} shows an example of detecting multiple patterns in an image, where multiple matched filters are employed which correspond to each pattern. 

\section{A Matched Filtering Perspective of CNN Operation}
\begin{figure}
	\centering
	\includegraphics[width=0.8\textwidth]{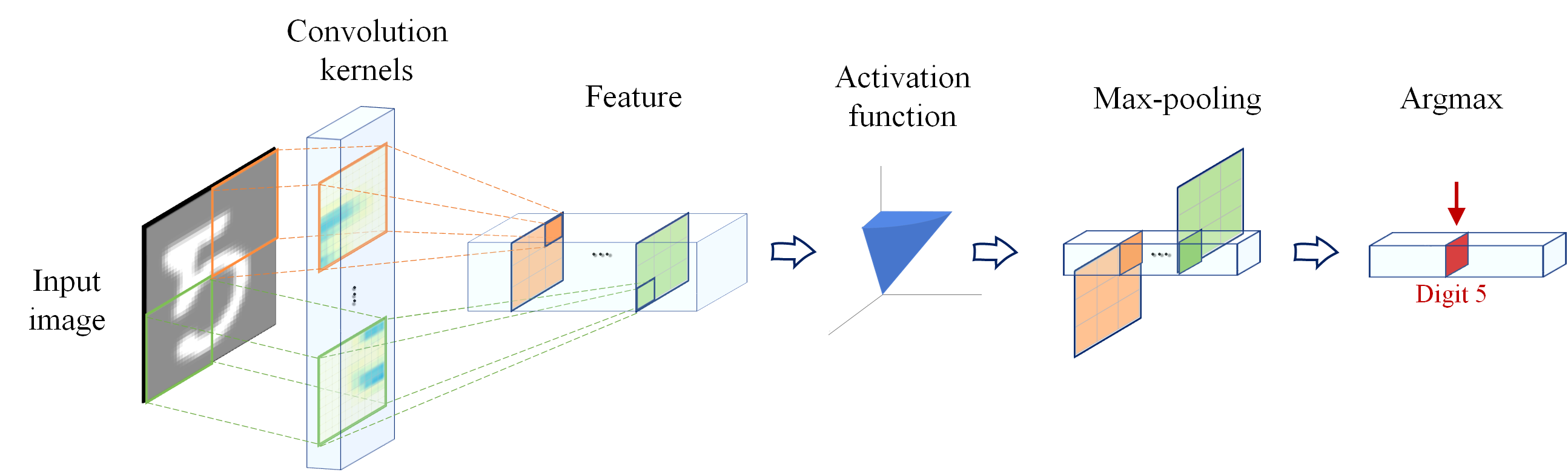}
	\vspace{-.5em}
	\caption{A basic pipeline of the CNN network for hand-written digit recognition, which consists of the convolution kernels, matched feature, non-linear activation function, max-pooling and one-hot (argmax) operations. The kernels are optimised, through adaptive training, to detect certain representative features by sliding over the whole input image, where the max-pooling and argmax operations are employed to identity the best match. }\label{fig_CNN_netstructure}
\end{figure}
\begin{figure}
\centering
\subfigure[Network architecture]{\includegraphics[width=0.65\textwidth]{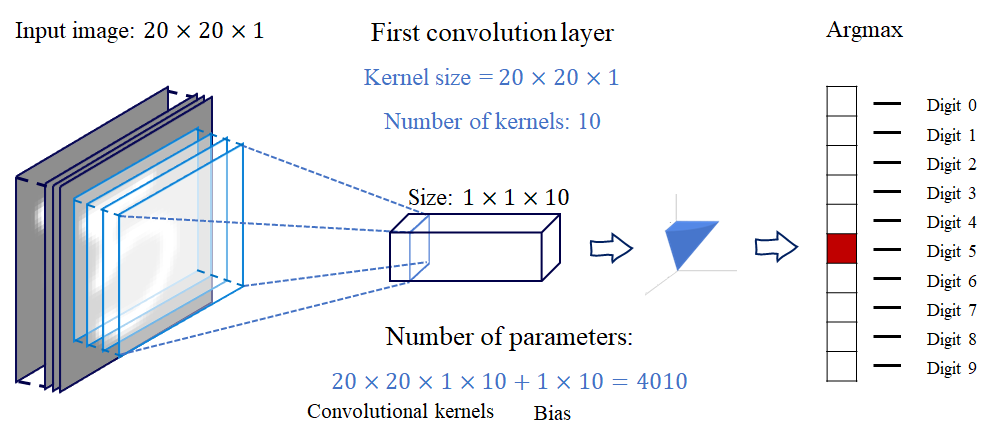}}
\subfigure[Pre-defined matched filters]{\includegraphics[width=0.35\textwidth]{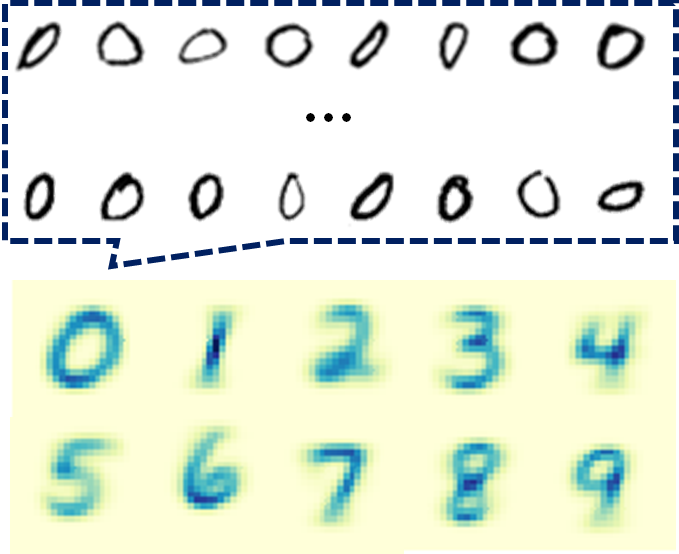}}
\subfigure[Matched filtering result]{\includegraphics[width=0.35\textwidth]{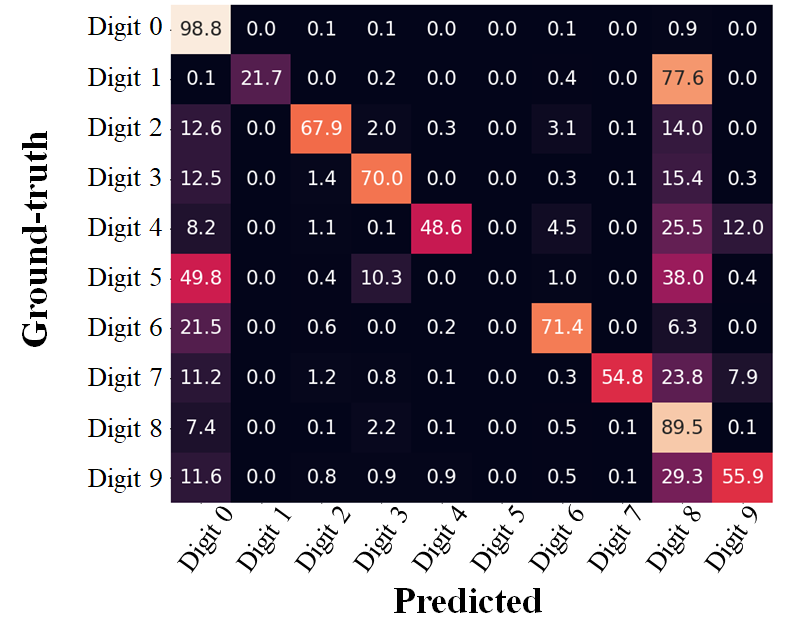}}
\caption{Illustration of a one-layer CNN that performs the matched filtering operation when recognising MNIST digits. The CNN network architecture is shown in (a). Because the CNN network has to recognise $10$ patterns (digits), $10$ pre-defined matched filters were employed which corresponds to the $10$ digits. Each filter was calculated as the average value of all the training images belonging to one digit class, which was then used to initialise  one convolutional kernel. Therefore, the CNN network consists of one convolution layer with $10$ convolutional kernels, and each kernel is of the size $20\times20$, the same size as the input images. The ReLU activation function was applied following the convolution layer so that the classification was performed by the one-hot operation. The pre-defined kernels are shown in (b). The matched filtering result on the test images is provided in (c), whereby the average accuracy is 57\%. }\label{fig_ini_filter_no_training}
\end{figure}
In matched filtering, the cross-correlation with a template plays the crucial role in detecting patterns with optimised SNR performances, and the patterns are located by maximising the filtered signal. In fact, as depicted in Fig. \ref{fig_CNN_netstructure}, the fundamentals of CNNs are almost the same as those in matched filtering, whereby in CNNs the template is known as the \textit{kernel} and the sliding is performed by a \textit{stride}. More specifically, each layer in CNNs typically has multiple kernels to detect different (parts of) patterns, which then effectively corresponds to a bank of multiple matched filters, as in \eqref{eq_matched_filter_multiple}. In this way, the convolution operation in CNNs can be formulated as
\begin{equation}\label{eq_cnn_mf}
    g_k(m, n) = (s*h_k)(m,n) = \sum_{a,b} s(a+c\cdot m,b+c\cdot n) \cdot f_k(a,b),
\end{equation}
where $f_k(\cdot,\cdot)$ represents the $k$-th kernel and $c$ denotes the stride step in the convolution operation. Then, a non-linear activation function is typically applied element-wise following the convolution operation in CNNs, with the aim to introduce the non-linearity so as to enhance generalisation capability. Among the extensive number of choices, such as the sigmoid \cite{mitchell1990machine}, tanh \cite{lecun2015deep} and softmax \cite{bridle1989training} non-linearities, the rectified linear unit (ReLU) activation function is one of the widely used activate functions \cite{agarap2018deep}, and is defined by 
\begin{equation}\label{eq_cnn_relu}
    \sigma_k(m,n) = \max\{0,g_k(m,n)\}.
\end{equation}
In high-level tasks such as object detection, classification and recognition, the decision is oftentimes made by \textit{predicting the best match}, which is achieved by the \textit{max pooling} or \textit{one-hot} operations; this can be formulated as 
\begin{equation}\label{eq_cnn_maxpooling}
	\begin{aligned}
    &\mathrm{max~pooling: ~} o(k)=\max_{m,n}\{\sigma_k(m,n)\} \\
    &\mathrm{one~hot: ~} k^*=\argmax_{k}\{o(k)\}. 
    \end{aligned}
\end{equation}
In this way, the positive prediction can be made when the maximal value (the best prediction) exceeds a pre-defined threshold. For example, in Fig. \ref{fig_single_asymfilters}-(a), when the threshold is set to $4$, the L-shaped pattern can be detected and located in the noisy image because we have $\max\{g(m,n)\}= g(1,1) = 5 > 4$. Otherwise, the negative prediction can be made. 

We further illustrate the way CNNs recognise digits and letters by performing matched filtering in two representative experiments. More specifically, we employ CNNs to recognise digits in the MNIST handwritten digit dataset \cite{lecun1998gradient}. The digit images were in grey-scale, which is convenient to exemplify convolutions in the 2D scenario. More importantly, since our goal is to recognise the digit given its various handwritten forms, the straightforward choice of the multiple matched filters, i.e., the kernels in CNNs, is the average value of images for each digit as shown in Fig. \ref{fig_ini_filter_no_training}-(a), and the handwritten images are regarded as certain noisy variants of the average image. As a consequence, the accuracy can approximate 60\% even without training the network, and the confusion matrix is provided in Fig. \ref{fig_ini_filter_no_training}-(b). 

\begin{figure}
\centering
\subfigure[Optimal filters from pre-defined kernels]{\includegraphics[width=0.35\textwidth]{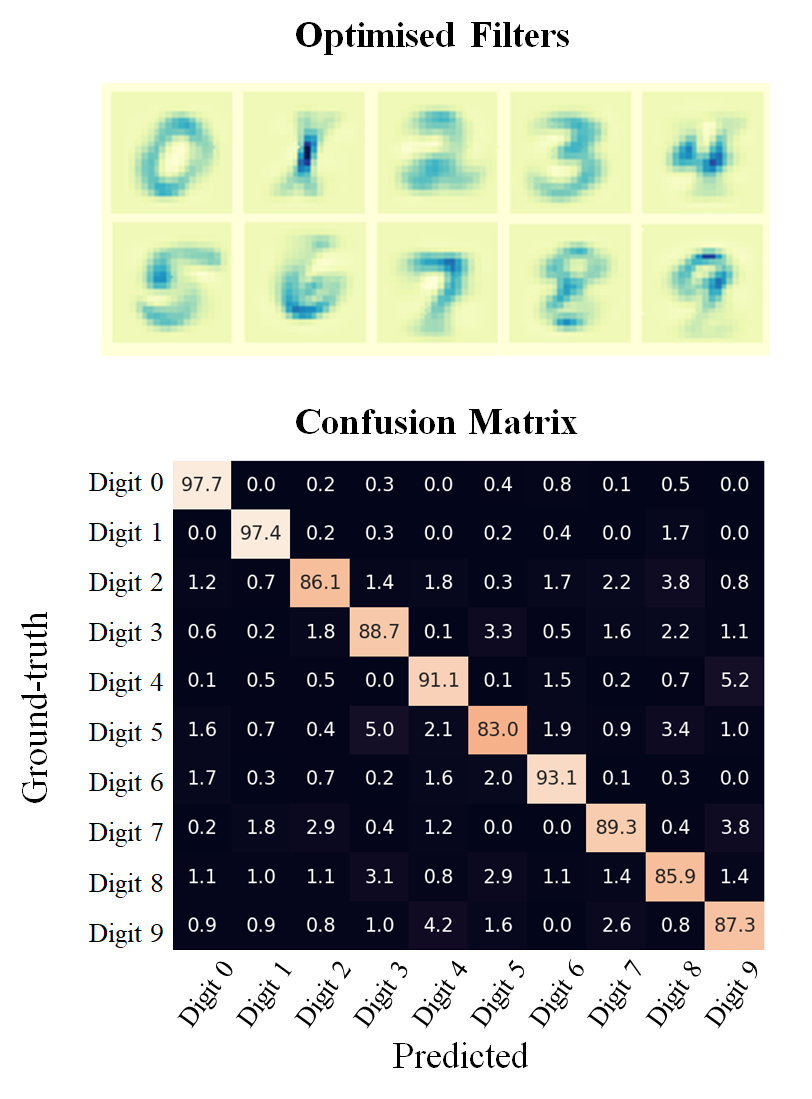}}
\subfigure[Optimal filters from random initialisation]{\includegraphics[width=0.35\textwidth]{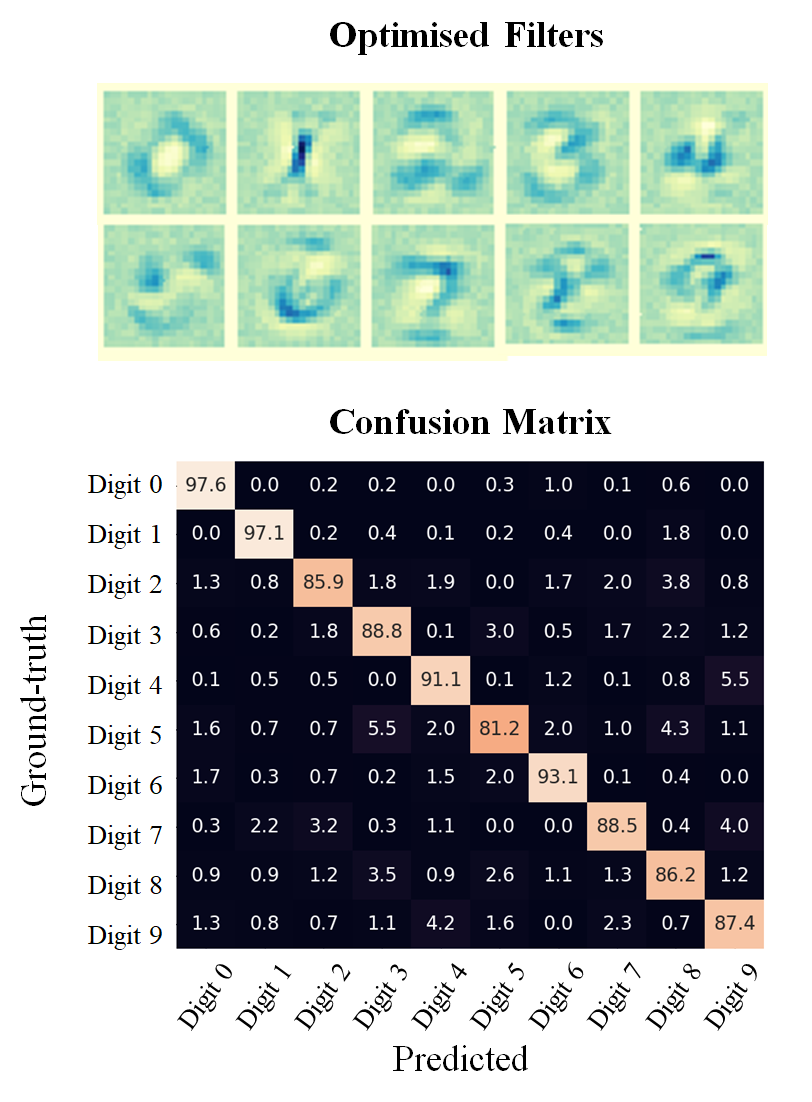}}
\caption{Illustration of training the one-layer CNN network for MNIST digits recognition, based on pre-defined and randomly initialised kernels. The pre-defined kernels are provided in Fig. \ref{fig_ini_filter_no_training}. Moreover, both kernels were further optimised by stochastic gradient descent over $100$ epochs, with the learning rate of $0.001$, until sufficient convergence. The average accuracy when optimising from pre-defined kernels in (a) is 90\%, whereas the average accuracy for random initialisation in (b) is 89\% on the test images. }\label{fig_matched_filter_trainingMNIST}
\end{figure}

\begin{figure}
\centering
\subfigure[Network architecture]{\includegraphics[width=0.8\textwidth]{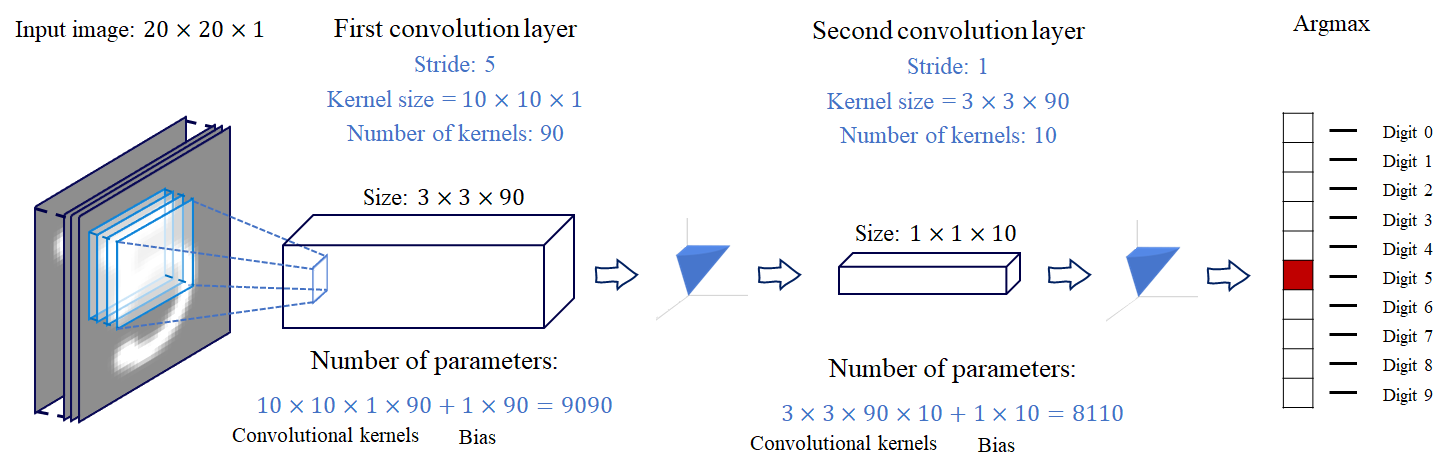}}
\subfigure[Matched filtering results]{\includegraphics[width=0.99\textwidth]{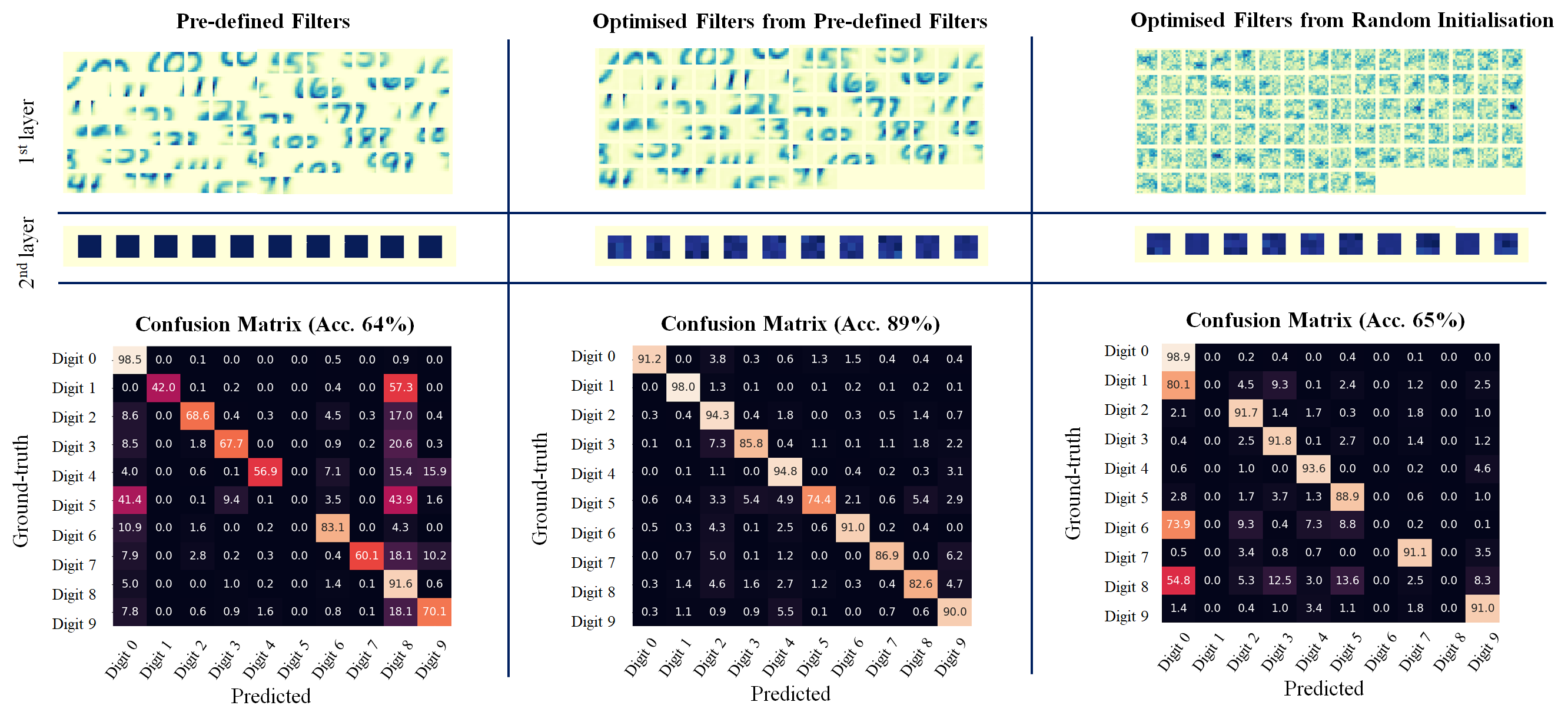}}
\caption{Illustration of training the CNN with two convolution layers for the recognition of MNIST digits. The network architecture is shown in (a). In this CNN, the kernel sizes were smaller than those of the one-layer CNN in Fig. \ref{fig_ini_filter_no_training}-(a). The matched filtering results are provided in (b). The pre-defined filters were calculated by the average local patches (small patterns) for each digit. In the second and third columns, the network was sufficiently optimised by stochastic gradient descent over $100$ epochs, with the learning rate of $0.001$. The Acc. in (b) denotes the average accuracy on the test images. }\label{fig_matched_filter_trainingMNIST_smallkernels}
\end{figure}

\begin{figure}
	\centering
	\includegraphics[width=0.99\textwidth]{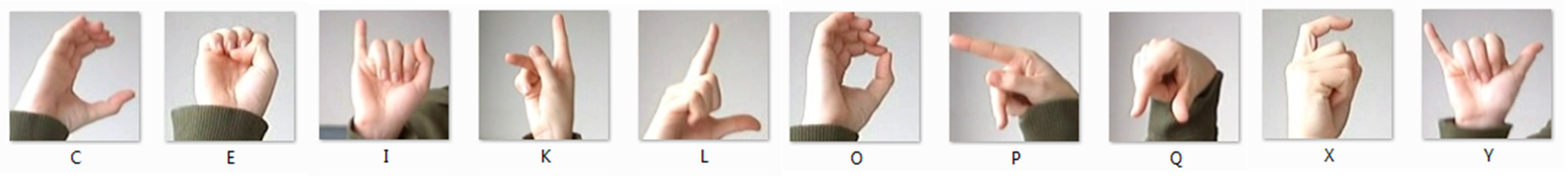}
	\vspace{-.5em}
	\caption{Examples of sign language images, where signs for letters \textit{C, E, I, K, L, O, P, Q, X, Y} were chosen. More specifically, the sign language dataset consists of $24$ letters and is available at \url{https://github.com/mon95/Sign-Language-and-Static-gesture-recognition-using-sklearn}. For better illustrations, we chose $10$ letters, the same number of classes as that of the MNIST dataset in our previous experiments. In the experiments for recognising sign language,  colour images were employed, which is a common practice in extensive applications. }\label{fig_sign_examples}
\end{figure}

\begin{figure}
	\centering
	\subfigure[Network architecture]{\includegraphics[width=0.8\textwidth]{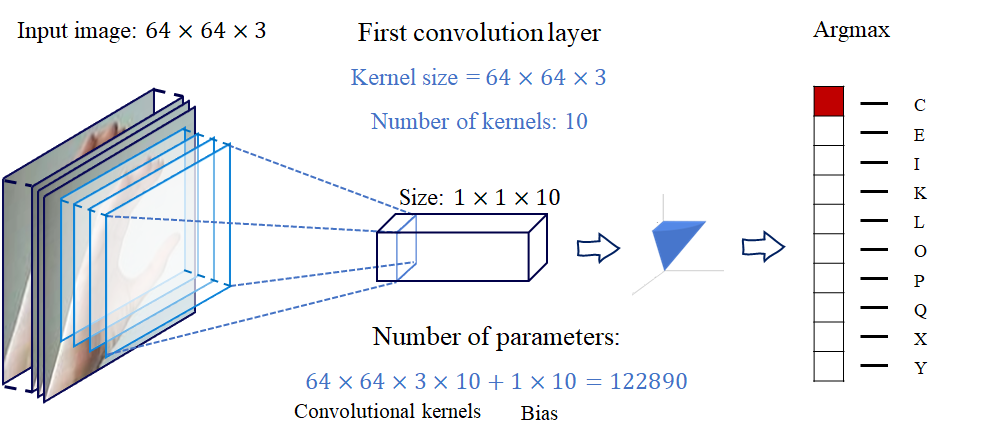}}
	\subfigure[Matched filtering results]{\includegraphics[width=0.99\textwidth]{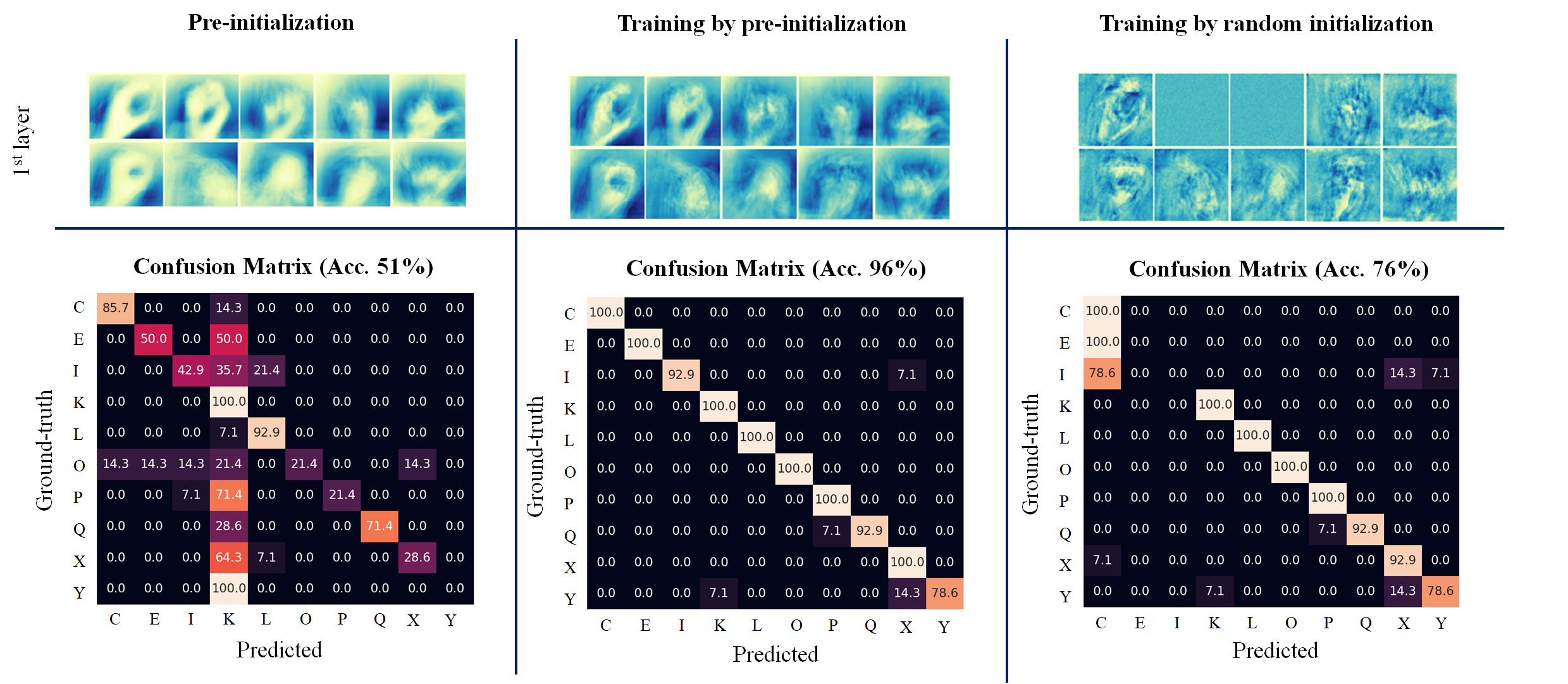}}
	\vspace{-.5em}
	\caption{Illustration of training the CNN with a single convolution layer when recognising sign language images. The network architecture is shown in (a). The matched filtering results are provided in (b). The pre-defined filters were calculated as the average value for each sign letter. In the second and third columns, the network was sufficiently optimised by stochastic gradient descent method over $100$ epochs, with the learning rate of $0.001$. Please note that Acc. in (b) denotes the average accuracy on the test images.}\label{fig_sign_one_layer}
\end{figure}

\begin{figure}
	\centering
	\subfigure[Network architecture]{\includegraphics[width=0.8\textwidth]{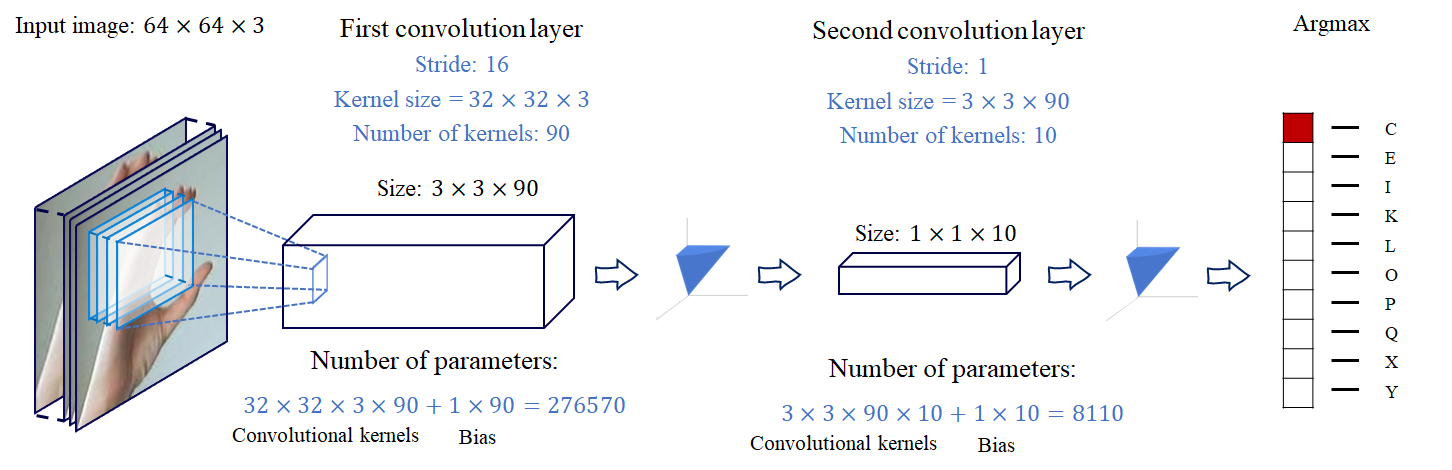}}
	\subfigure[Matched filtering results]{\includegraphics[width=0.99\textwidth]{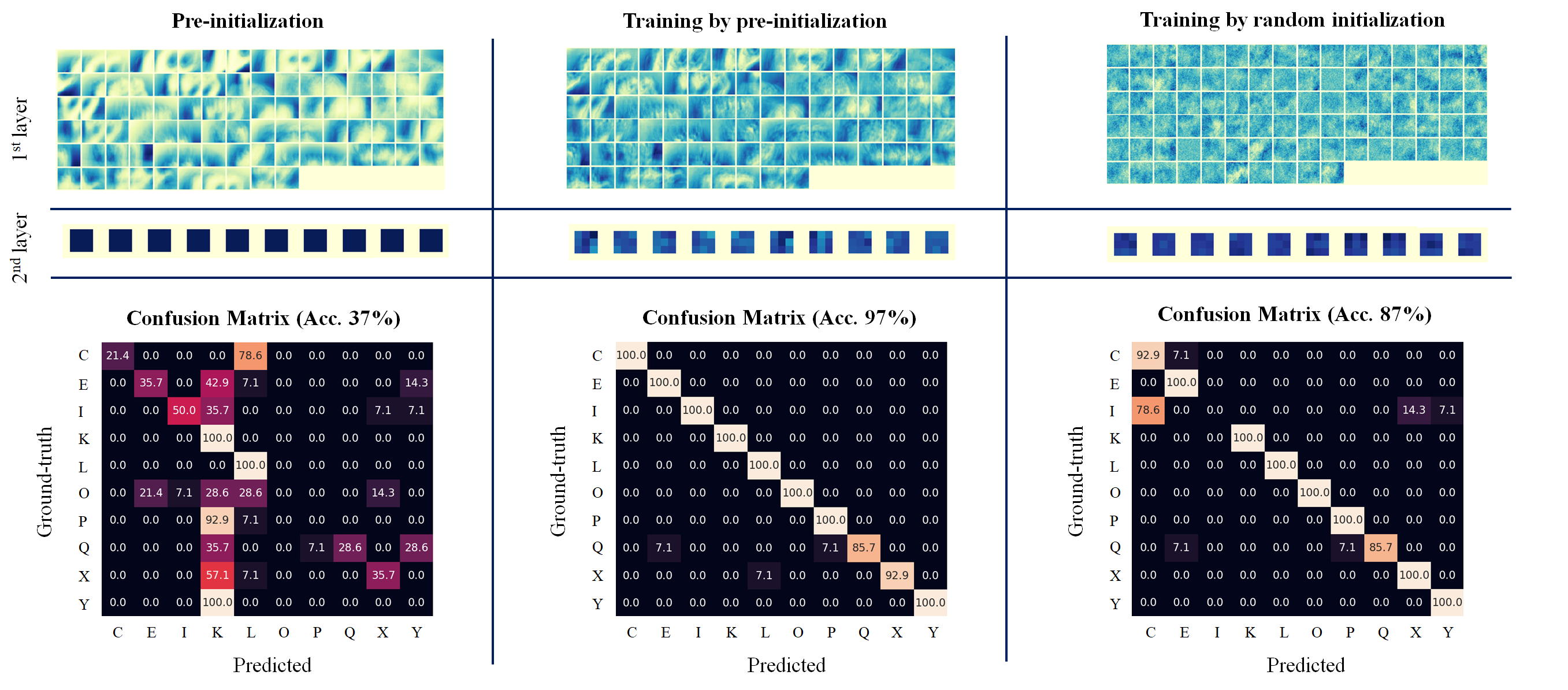}}
	\vspace{-.5em}
	\caption{Illustration of training the CNN with two convolution layers when recognising sign language images. The network architecture is shown in (a). The matched filtering results are provided in (b). The pre-defined filters were calculated by the average local patches (small patterns) for each sign letter. In the second and third columns, the network was sufficiently optimised by stochastic gradient descent over $100$ epochs, with the learning rate of $0.001$. Please note that Acc. in (b) denotes the average accuracy on the test images.}\label{fig_sign_two_layers}
\end{figure}

Moreover, it can be seen from Fig. \ref{fig_ini_filter_no_training}-(b) that the main error lies in predicting digit 8. This might be due to the fact that the pre-defined kernel for the digit 8 highly correlates with other digits, which violate the assumption of independent additive noise. To address this issue, we can adaptively adjust the matched filters so as to improve the recognition accuracy. In other words, the matched filters can be optimised by the \textit{back-propagation} technique in CNNs. The so-adjusted filters are illustrated in Fig. \ref{fig_matched_filter_trainingMNIST}. It is obvious that by optimising the matched filters,  prediction accuracy significantly improves, from 57\% in Fig. \ref{fig_ini_filter_no_training}-(c) to 90\% in Fig. \ref{fig_matched_filter_trainingMNIST}. More importantly, optimising the matched filters in the CNN network with a pre-defined content (i.e., Fig. \ref{fig_matched_filter_trainingMNIST}-(a)) contributes to the average accuracy, compared to the widely employed random initialisation (i.e., Fig. \ref{fig_matched_filter_trainingMNIST}-(b)). Furthermore, we illustrate in Fig. \ref{fig_matched_filter_trainingMNIST_smallkernels} that the gain from the pre-defined kernels became more significant for smaller kernel sizes, a commonplace in practical CNNs. 
To further illustrate the principle of matched filtering colour images by CNNs, we designed filters based on recognising sign language letters, as shown in  Fig. \ref{fig_sign_examples}. The optimisation and recognition by the one-layer CNN is given in Fig. \ref{fig_sign_one_layer}, whilst the results for the CNN with two convolution layers are given in Fig. \ref{fig_sign_two_layers}. 

%\section{Optimal Matched Filters}
\section{conclusion}
We have introduced the basic principles of CNNs for images, from the perspective of matched filtering. In particular, whilst enjoying great success, CNNs still suffer from their \textit{black-box} nature that prevents their theoretical analysis and physical meaning related to their learnt parameters. We have addressed this issue by justifying the use of convolution operation in CNNs, the basic building blocks of deep CNNs, from the perspective of matched filtering in signal processing. In this way, the convolution kernels in CNNs have found their root in  matched filters. Next, the convolution-activation-pooling chain has been theoretically supported by the maximal SNR performance of matched filtering, such that the basic operations of CNNs can be theoretically analysed under the well understood umbrella of matched filtering. To further illustrate the connection between CNNs and matched filtering, we have provided extensive examples, including both synthetic and real-world experiments, whereby the physical meaning of learnt CNNs has been revealed. It is our hope that by connecting CNNs to signal matched filtering will provide a new way to demystify CNNs, together with paving the way for applying signal processing methods in constructing new architectures, and establishing new theories of CNNs.
\bibliographystyle{IEEEtran}
\bibliography{IEEEfull,ShengxiLi}
\end{document}